%% file: main.tex
\documentclass[runningheads]{llncs}
\usepackage[T1]{fontenc}
\usepackage{amssymb}
\usepackage{amsmath}
 \usepackage{multirow}
 \usepackage{tikz}
\usepackage{graphicx}
\usepackage{booktabs}

\begin{document}
\title{Uncertainty-aware Language Guidance for Concept Bottleneck Models}
%
%
%
\author{Yangyi Li\inst{1}\ \and
Mengdi Huai\inst{1}}
\authorrunning{Y. Li et al.}
%
\institute{Iowa State University, Ames IA 50011, USA \\
\email{\{liyangyi, mdhuai\}@iastate.edu}}
\maketitle              
\begin{abstract}
Concept Bottleneck Models (CBMs) provide inherent interpretability by first mapping input samples to high-level semantic concepts, followed by a combination of these concepts for the final classification. However, the annotation of human-understandable concepts requires extensive expert knowledge and labor, constraining the broad adoption of CBMs. On the other hand, there are a few works that leverage the knowledge of large language models (LLMs) to construct concept bottlenecks. Nevertheless, they face two essential limitations: First, they overlook the uncertainty associated with the concepts annotated by LLMs and lack a valid mechanism to quantify uncertainty about the annotated concepts, increasing the risk of errors due to hallucinations from LLMs. Additionally, they fail to incorporate the uncertainty associated with these annotations into the learning process for concept bottleneck models. To address these limitations, we propose a novel uncertainty-aware CBM method, which not only rigorously quantifies the uncertainty of LLM-annotated concept labels with valid and distribution-free guarantees, but also incorporates quantified concept uncertainty into the CBM training procedure to account for varying levels of reliability across LLM-annotated concepts. We also provide the theoretical analysis for our proposed method. Extensive experiments on the real-world datasets validate the desired properties of our proposed methods. 
\vspace{-2mm}

\end{abstract}

\input{secs/1_intro}

\input{secs/2_related_work}
\input{secs/4_method}

\input{secs/5_experiment}
\input{secs/6_conclusion}

\vspace{-3mm}
\bibliographystyle{splncs04}
\bibliography{reference}

\end{document}

%% file: secs/1_intro.tex
\vspace{-8mm}
\section{Introduction }
\label{sec:intro}
\vspace{-4mm}
The impressive capabilities of deep neural networks (DNNs) have led to their successful deployment across a wide range of applications, but this success often comes at the expense of transparency, rendering them ``black boxes''~\cite{sun2020cracking}. This opacity hinders their widespread adoption in high-stakes domains where the rationale behind a decision is crucial for both trust and accountability. In response to this need for inherent explainability, Concept Bottleneck Models (CBMs) have emerged as a promising solution~\cite{koh2020concept,yuksekgonul2022post}. CBMs integrate interpretability directly into their architecture by introducing an intermediate concept layer that forces the model to base its predictions on a set of human-understandable attributes, thus offering a transparent decision-making process.

Despite their promise, the practical deployment of CBMs is often hindered by their reliance on concept annotations provided by human experts. This manual process is labor-intensive, costly, and a significant bottleneck to scalability. To overcome this, there are a few works that adopt Large Language Models (LLMs)~\cite{ouyang2022training,ren2024survey,han-etal-2025-attributes,qian2025towardsben} to automate concepts~\cite{srivastava2024vlg,yang2023language}. However, they suffer from two essential limitations. First, they overlook the uncertainty associated with the concepts annotated by LLMs, thereby increasing the risk of errors caused by hallucinations. These methods directly rely on LLMs to produce concept labels. However, due to the inherent possibility of hallucinations~\cite{snyder2024early}, irrelevant and imprecise concepts may be generated, and the resulting concept annotations may be unreliable. Such noisy and unreliable concept labels can propagate through the CBM pipeline, degrading model reliability and interpretability.

Additionally, when using large language models to annotate concepts in the training data, existing approaches fail to incorporate the uncertainty associated with these annotations into the learning process for concept bottleneck models. To train CBMs, they simply adopt the deterministic concept labels output by large language models, without modeling the confidence or variability of these labels. However, uncertainty-aware concept annotations inherently convey richer supervisory information, since they capture the varying degrees of unreliability in the concepts predicted by LLMs. When such uncertainty is ignored and annotations are reduced to deterministic labels, much of this informative signal is lost, which can ultimately hinder the effectiveness of CBM training.

However, addressing these limitations presents several unique research challenges. First, it is inherently difficult to quantify the uncertainty of concepts annotated by LLMs in a principled manner with guarantees. Traditional uncertainty quantification methods~\cite{lakshminarayanan2017simple,qian2025towards} often assume that data is independent and identically distributed (i.i.d.), but such an assumption is difficult to justify for the LLM-based annotation process across concepts and inputs. Furthermore, prior works involving language-guided CBMs typically resort to simple and non-adaptive confidence thresholds without providing theoretical coverage guarantees. They fail to rigorously account for the hallucinations inherent in LLMs, leading to unreliable concept bottlenecks where irrelevant or imprecise concepts may propagate errors to the final prediction. Second, it is challenging to effectively incorporate quantified concept uncertainty into the CBM training pipeline. After leveraging LLMs to generate uncertainty-aware concept annotations, the training samples for specific generated concepts may become sparse based on varying reliability levels. If CBMs are trained directly on such data, the model tends to disregard the informative signals from these sparse uncertainty-aware concepts. Consequently, the CBM fails to fully learn the special knowledge embedded in these concepts, limiting the integration of uncertainty information and ultimately hindering improvements in model performance and interpretability.

To bridge this gap, in this paper, we propose \textbf{U}ncertainty-aware \textbf{L}anguage Guidance for \textbf{C}oncept \textbf{B}ottleneck \textbf{M}odels (\textbf{ULCBM}), which can not only rigorously quantify the uncertainty of LLM-annotated concepts with distribution-free guarantees but also effectively incorporate quantified concept uncertainty into the CBM training procedure to account for varying levels of reliability across LLM-annotated concepts.
Specifically, we first propose a novel uncertainty quantification method capable of evaluating the confidence of each annotated concept across three complementary dimensions. To address the challenges of distribution assumptions and the absence of theoretical guarantees, we propose to leverage Conformal Prediction (CP)~\cite{vovk2005algorithmic,shafer2008tutorial,li2024data,chen2024modeling,zhao2025membership} to calibrate a global acceptance threshold on a calibration set. This framework enables us to construct concept sets that provably satisfy user-specified risk levels over exchangeable samples, without the i.i.d. assumption on the concept annotation mechanism. We quantify these dimensions through three corresponding loss functions, including discriminability to measure the specificity of concepts to the image's true class, coverage to ensure the selected subset represents the full semantic scope of candidate concepts, and diversity to penalize semantic redundancy among selected attributes.
Additionally, based on the quantified uncertainty, we design a targeted data augmentation pipeline to integrate uncertainty information into the learning process. To address the challenge of sparse training samples caused by varying reliability levels, we propose to synthesize additional examples for specific sparse concepts by inserting reliable patches from source images into target samples. Crucially, this insertion process is strictly guided by the uncertainties derived from our defined criteria, ensuring the placement of new patches avoids overlapping with existing concepts chosen by our uncertainty threshold. This strategy effectively mitigates the data scarcity of high-reliability but rare concepts, ensuring the CBM does not overlook these informative signals. We also provide a theoretical analysis proving that our calibration procedure ensures the expected losses for these criteria satisfy their prescribed risk levels with distribution-free guarantees. Extensive experiments on real-world datasets validate the desired properties and superior performance of our proposed methods.

%% file: secs/2_related_work.tex
\vspace{-4mm}
\section{Related Work}
\label{sec:related}
\vspace{-3mm}
Concept Bottleneck Models (CBMs) are a class of self-explaining neural networks that force predictions through an intermediate layer of human-understandable concepts~\cite{koh2020concept,oikarinen2023label,qian2024towards}. Importantly, CBMs play a pivotal role in various applications, including model debugging \cite{ismail2023concept}, object tracking \cite{pittino2023hierarchical}, and human intervention on decisions \cite{chauhan2023interactive}. For example, in the context of human intervention, experts can manually correct mispredicted concepts (e.g., changing ``wing color'' from blue to red) at test time to steer the model toward the accurate class label. However, acquiring expert annotations for human-understandable concepts is expensive, which severely restricts the scalability and deployment of CBMs in practice. To address this, recent works~\cite{srivastava2024vlg,yang2023language} have attempted to leverage LLMs to automatically generate useful concepts, but they face several essential limitations. First, they fail to rigorously quantify the inherent uncertainty of LLM-annotated concepts with theoretical coverage guarantees. Therefore, they cannot effectively prevent the generation of irrelevant and imprecise concepts due to the inherent possibility of hallucinations. Furthermore, the training procedures in existing works generally treat these generated concepts as deterministic ground truths, thereby neglecting the valuable supervisory signal inherent in their quantified uncertainty. In this work, we introduce a framework that addresses these limitations by deriving concept uncertainty with valid and distribution-free guarantees, and integrating this quantified uncertainty directly into the CBM training pipeline to enhance both model reliability and training effectiveness.

%% file: secs/4_method.tex
\vspace{-3mm}
\section{Methodology}
\vspace{-3mm}
\label{sec:method}

In this section, we introduce our proposed framework called ULCBM. While automating CBM annotation with LLMs is promising, current methods lack mechanisms to quantify uncertainty with theoretical guarantees and neglect the rich supervisory information for specific uncertainty-aware concepts. ULCBM addresses these limitations by integrating a principled uncertainty quantification framework with a targeted data augmentation pipeline. We first detail the uncertainty-aware concept generation with language guidance, which provides distribution-free uncertainty guarantees across three complementary dimensions. Based on the derived concept uncertainties, we then describe the process for using a targeted data augmentation pipeline to synthesize extra training samples of specific concepts, facilitating the effective training of uncertainty-aware CBMs.

\vspace{-5mm}
\subsection{Uncertainty-aware Generation with Language Guidance}
\vspace{-2mm}

We now detail the uncertainty-aware concept generation process with language guidance, designed to construct concept annotations endowed with formal and distribution-free guarantees. Let $\mathcal{X} = \mathbb{R}^{H \times W \times 3}$ denote the input image space and $\mathcal{Y} = \{1, 2, \ldots, L\}$ denote the label space, where $L$ is the number of classes. Denote $D^{\text{tr}}  = \{(x_i, y_i)\}_{i=1}^{N^{\text{tr}}}$, $x_i \in \mathcal{X}$, $y_i \in \mathcal{Y}$ as the training dataset, where $x_i$ is the $i$-th image and $y_i$ is the corresponding label. Let $\mathcal{S}$ be a set of fine-grained and natural language concepts. We propose to generate a modified dataset $\hat{D}^{\text{tr}}$ from $D^{\text{tr}}$. In $\hat{D}^{\text{tr}}$, each image $x_i$ is annotated with not only its original class label $y_i$ but also a set of concepts deemed useful for prediction.

First, we use an LLM to generate candidate concepts $S_l$ for each class $l \in \mathcal{Y}$ following the prompting strategy of~\cite{oikarinen2023label}. For example,  we adopt the prompt: ``List the most important features for recognizing something as a `goldfish'.'' We then employ Grounding-DINO~\cite{liu2024grounding} with a Swin-B backbone, a leading grounded object detector, to obtain bounding boxes for these candidate concepts within the dataset. For each image $x_i$ with class label $l$ and candidate concepts $S_l$, we prompt the Grounding-DINO model with $S_l$ and obtain $K_i$ bounding boxes $B_i = \{(b_j, t_j, s_j)\}_{j=1}^{K_i}$, where $b_j \in \mathbb{R}^{4 \times 2}$ is the $j$-th bounding box coordinates, $t_j \in \mathbb{R}$ is the corresponding confidence given by the model and $s_j \in S_l$ is the concept associated with that bounding box. However, this raw generation process cannot quantify uncertainty with distribution-free guarantees, leaving the annotations vulnerable to hallucinations regarding whether each concept is truly present. 

To overcome this limitation, we integrate uncertainty quantification with distribution-free guarantees into our concept generation pipeline, systematically identifying reliable concepts across three complementary dimensions. Let \(D^{\text{cal}} = \{(x_i,y_i)\}_{i=1}^{N^{\text{cal}}}\) denote the calibration data. Then, for \(x_i \in D^{\text{tr}}\), we construct its uncertainty concept set as $ \mathcal{C}_\lambda(x_i;B_i) = \{s_j: (b_j,t_j,s_j) \in B_i, t_j  \geq1-\lambda\}$, where $\lambda \in [0,1]$ is a parameter that increases the size of the prediction sets as its value grows.
We evaluate the quality of the uncertainty set \(\mathcal{C}_\lambda(x_i;B_i)\) based on three criteria: discriminability, coverage, and diversity, quantified via corresponding loss functions. These three losses are designed to be complementary: discriminability ensures the selected concepts are relevant to the correct class, coverage ensures they are comprehensive enough to represent the class, and diversity ensures they are not redundant. Together, they guide the construction of a final concept set \(\mathcal{C}_\lambda(x_i;B_i)\) that is relevant, representative, and non-redundant.

We first introduce a discriminability loss to ensure that the selected concepts are substantially more aligned with the image  $x_i$ than concepts associated with other classes. We first quantify the similarity between an image $x_i$ and a concept $s$ using the cosine similarity $
\text{Sim}(x_i, s) = 1+\text{cos}(\mathcal{T}_m(x_i), \mathcal{T}_t(s))$,
where $\mathcal{T}_m$ is the image encoder and $\mathcal{T}_t$ is the text encoder. The discriminability loss $\ell_{\text{dis}}$ is then defined as a ratio measuring the collective similarity of the concepts in $\mathcal{C}_\lambda(x_i;B_i)$ to the image $x_i$, relative to the similarity of all concepts from competing classes:
\vspace{-5mm}
\begin{align}
\ell_{\text{dis}}(\mathcal{C}_\lambda(x_i;B_i), x_i, y_i) = 1-\frac{ \sum_{s \in \mathcal{C}_\lambda(x_i;B_i)} \text{Sim}(x_i, s)}
{\sum_{y \neq y_i} \sum_{s \in S_{y}} \text{Sim}(x_i, s)}.
\end{align}
\vspace{-4mm}

\noindent Minimizing $\ell_{\text{dis}}(\mathcal{C}_\lambda(x_i;B_i), x_i, y_i)$ encourages the construction of concept sets that are highly specific and relevant to the image's true class \(y_i\).

Next, we denote the cosine dissimilarity between the features of the two concepts extracted by the text encoder as $\varphi(s_1, s_2) = \frac{1-\cos( \mathcal{T}_t(s_1), \mathcal{T}_t(s_2) )}{2}$,
where lower values of $\varphi(s_1, s_2)$ correspond to higher similarity. The coverage loss is then defined as the average dissimilarity from each concept in the full candidate set $S{y_i}$ to its nearest neighbor within the constructed set:
\vspace{-2.6mm}
\begin{align}
\ell_\text{cov}(\mathcal{C}_\lambda(x_i;B_i),x_i, y_i) = \frac{1}{|S_{y_i}|}\sum\limits_{s_1 \in S_{y_i}} \min\limits_{s_2 \in \mathcal{C}_\lambda(x_i;B_i)} \varphi(s_1, s_2).
\end{align}\vspace{-4mm}

\noindent Minimizing \(\ell_\text{cov}\) ensures that the selected concepts are not clustered in one part of the semantic space but instead provide good coverage over the entire set of candidate concepts $S_{y_i}$ defining the class \(y_i\).

To further encourage semantic diversity among selected concepts, we introduce a diversity loss $\ell_{\text{div}}$. This loss penalizes semantic redundancy by computing the total pairwise dissimilarity within the selected set $\mathcal{C}_\lambda(x_i;B_i)$, normalized by the corresponding sum over the full class-level concept pool $S_{y_i}$ containing all available candidate concepts. Formally, the loss is defined as:
\vspace{-2.6mm}
\begin{align}
\ell_{\text{div}}(\mathcal{C}_\lambda(x_i;B_i),x_i, y_i)
= 1 - 
\frac{\sum_{\{s_1,s_2\}\subseteq \mathcal{C}_\lambda(x_i;B_i)} \varphi(s_1,s_2)}
{\sum_{\{s_1,s_2\}\subseteq S_{y_i}} \varphi(s_1,s_2)}.
\end{align}
\vspace{-4mm}

\noindent Minimizing the loss $\ell_\text{div}$ promotes the selection of a semantically diverse set of concepts, thereby avoiding redundancy.

Note that if the concept set $\mathcal{C}_\lambda(x_i;B_i)$ is empty (or contains fewer than two concepts for $\ell_\text{div}$), we define the corresponding loss as 1, representing the maximum penalty. The three loss functions (e.g., \(\ell_{\text{dis}}, \ell_{\text{cov}}, \ell_{\text{div}}\)) all decrease as the set 
$\mathcal{C}_\lambda(x_i;B_i)$ is improved. However, a key goal for interpretability is to find the most concise concept sets possible. Thus, we aim to find the smallest concept sets that still satisfy user-specified risk levels $\alpha_\text{dis}, \alpha_\text{cov}, \alpha_\text{div}$. This goal is formalized by the following optimization problem, which seeks a threshold $\lambda$ to be applied to all samples drawn from the target data distribution:
\vspace{-2mm}
\begin{align}
\label{eq:Find}
    \min_\lambda&\mathbb{E}[|\mathcal{C}_\lambda(x_i; B_i)|] \quad \text{s.t.} \ \mathbb{E}[\ell_\text{dis}(\mathcal{C}_\lambda(x_i; B_i), x_i, y_i)] \leq \alpha_\text{dis}, \\
    &\mathbb{E}[\ell_\text{cov}(\mathcal{C}_\lambda(x_i; B_i) ,x_i, y_i)] \leq \alpha_\text{cov},\;\; \mathbb{E}[\ell_\text{div}(\mathcal{C}_\lambda(x_i; B_i),x_i, y_i)] \leq \alpha_\text{div}. \notag
\end{align}\vspace{-6mm}

\noindent Nonetheless, directly solving the optimization problem above is challenging. First, the objective function involving \(|\mathcal{C}_\lambda(x_i; B_i)|\) is discontinuous with respect to \(\lambda\) and is therefore non-differentiable. Second, the constraints are expectations over the data distribution. Consequently, directly optimizing on a finite sample offers no rigorous guarantee that the risk levels will hold on the whole data.

Therefore, instead of direct optimization, we turn to a practical approach that provides distribution-free uncertainty quantification with a formally theoretical guarantee. Concretely, we use the calibration set to select an appropriate \(\lambda\) that provably satisfies the required risk constraints and the objective about small set size. Specifically, for each loss, we compute an empirical risk on the calibration set. To illustrate, we consider the discriminability loss $\ell_{\text{dis}}$ and compute its empirical risk on the calibration set by $
\widehat{R}^{\text{dis}}_{\,N^{\text{cal}}}(\lambda)
=\frac1{N^{\text{cal}}}\sum_{i=1}^{N^{\text{cal}}}      \ell_{\text{dis}}(\mathcal{C}_\lambda(x_i;B_i),x_i,y_i)$, 
and define the smallest threshold that meets the desired risk budget $
\hat{\lambda}_{\text{dis}} = \inf\{\lambda: \widehat{R}^{\text{dis}}_{\,N^{\text{cal}}}(\lambda) \le\alpha_{\text{dis}}-\tfrac{1-\alpha_{\text{dis}}}{N^{\text{cal}}}\}.$ If this set is empty, we define $\hat{\lambda}_{\text{dis}} = 1$.
Because \(\widehat{R}^{\text{dis}}_{\,N^{\text{cal}}}(\lambda)\) is non‑increasing in \(\lambda\),
\(\hat{\lambda}_{\text{dis}}\) can be found efficiently via binary search.
Repeating the same procedure for \(\ell_{\text{cov}}\) and \(\ell_{\text{div}}\) yields
\(\hat{\lambda}_{\text{cov}}\) and \(\hat{\lambda}_{\text{div}}\).
Finally, to ensure all three constraints are met simultaneously, we adopt the most conservative value as $\hat{\lambda}= \max\{\hat{\lambda}_{\text{dis}}, \hat{\lambda}_{\text{cov}},
\hat{\lambda}_{\text{div}}\}$. Since each empirical risk $\widehat{R}^{k}_{N^{\text{cal}}}(\lambda)$ is non-increasing in $\lambda$ for each $k \in \{\text{dis}, \text{cov}, \text{div}\}$, choosing $\hat{\lambda}_k$ as the smallest value that meets the risk budget automatically produces the smallest admissible concept sets for each loss. Taking $\hat{\lambda} = \max_k \hat{\lambda}_k$ then selects the smallest global threshold that satisfies all three constraints simultaneously, which matches the objective in Eq.~\eqref{eq:Find} of minimizing set size under the desired bounds.

\vspace{-2mm}
\begin{theorem}
\label{thm:CRCThm} 
Assume that the calibration set $D^{\text{cal}}$ and the target sample $(x_i, y_i)$ are exchangeable. For any desired risk level $\alpha_\text{dis}, \alpha_\text{cov}, \alpha_\text{div}\in (0,1)$, we obtain the thresholds $\hat{\lambda}_k$ for each $k \in \{\text{dis}, \text{cov}, \text{div}\}$ by computing
$ \hat{\lambda}_{k} = \inf\{\lambda:\, \widehat{R}^{k}_{\,N^{\text{cal}}}(\lambda) \le\alpha_{k}-\tfrac{1-\alpha_{k}}{N^{\text{cal}}}\}$,
and denote $\hat{\lambda} = \max\{\hat{\lambda}_\text{dis},\hat{\lambda}_\text{cov},\hat{\lambda}_\text{div}\}$. Then, the constructed uncertainty set \(\mathcal{C}_{\hat{\lambda}}(x_i;B_i)\)  satisfies $\mathbb{E}[\ell_k(\mathcal{C}_{\hat{\lambda}}(x_i; B_i) ,x_i, y_i)]\le \alpha_k$ in Eq.~\eqref{eq:Find} for each $k$, where the expectation is taken over the distribution of $(x_i,y_i)$.
\end{theorem}
\vspace{-2mm}
For the constructed uncertainty set $\mathcal{C}_{\hat{\lambda}}(x_i;B_i)$, Theorem~\ref{thm:CRCThm} guarantees that the expected value of each loss remains below its desired risk level. Concretely, under exchangeability, the calibration losses $\ell_k(\mathcal{C}_{\lambda}(x_i;B_i))$ computed on $D^{\text{cal}}$ are exchangeable with the corresponding loss evaluated on a target sample. Conformal risk control~\cite{angelopoulos2022conformal,li-huai-2025-quantifying} then selects $\hat{\lambda}_k$ using a calibration-based one-sided bound (with the finite-sample correction term $\tfrac{1-\alpha_k}{N^{\text{cal}}}$), ensuring $\mathbb{E} [\ell_k (\mathcal{C}_{\hat{\lambda}_k}(x_i;B_i) )]\le \alpha_k$ for each $k\in\{\text{dis},\text{cov},\text{div}\}$ and a target sample $(x_i, y_i)$ exchangeable with $D^{\text{cal}}$. Finally, taking $\hat{\lambda}=\max_k \hat{\lambda}_k$ yields simultaneous control over all three losses.

After determining the suitable \(\hat{\lambda}\) in Theorem~\ref{thm:CRCThm}, we construct the final global concept vocabulary $\hat{S}=\{ s \in S \mid s \in \bigcup_{i=1}^{|D^{\text{tr}}|} \mathcal{C}_{\hat{\lambda}}(x_i;B_i)\}$ by taking the union of all concepts that appear in at least one image's calibrated set. Using this vocabulary $\hat{S}$, the one-hot encoded concept label vector $o_i \in \{0,1\}^{|\hat{S}|}$ for image $x_i$ is thus defined as $(o_i)_j = \mathbf{1}(s_j \in \mathcal{C}_{\hat{\lambda}}(x_i;B_i))$.
Then we can obtain the final concept-labeled dataset $\hat{D}^{\text{tr}}= \{(x_i, o_i, y_i)\}_{i=1}^{|D^{\text{tr}}|}$ with guarantees for training CBMs.

\vspace{-4mm}
\subsection{Training Uncertainty-aware CBMs}
\vspace{-2mm}
As established previously, a critical challenge arises where the varying reliability levels of LLM-annotated concepts lead to sparse supervisory signals for specific concepts after rigorous filtering. To address this, we introduce a targeted data augmentation pipeline designed to synthesize additional valid training samples, thereby facilitating the effective training of uncertainty-aware CBMs.

The process begins by identifying a specific sparse concept $s^r$ whose reliable occurrences fall below a predefined threshold. For a given image $x_i$ of the associated class $l$, we insert a representative visual patch of $s^r$. Crucially, this insertion process is strictly guided by our derived uncertainty, where the pipeline samples a placement window $\hat{b}$ that avoids overlapping with any existing concept filtered by our uncertainty threshold $\hat{\lambda}$ (defined in Theorem~\ref{thm:CRCThm}). Formally, we define the set of reliable existing boxes as $B_i'=\{(b_j,t_j,s_j)\in B_i\mid t_j\geq 1-\hat{\lambda}, s_j \neq s^r\}$, and constrain the placement such that $\hat{b} \subseteq [0,W) \times [0,H)$ and $\hat{b} \cap b_j = \varnothing, \forall (b_j, t_j, s_j) \in B_i'$. 
To source the visual content, we randomly select another training image whose uncertainty concepts contain $s^r$ and crop a patch $(x_{s^r}^{\text{src}}, M_{s^r})$ from its annotated region. Here, $x_{s^r}^{\text{src}}$ represents the image pixels, and $M_{s^r}$ is the binary mask. $M_{s^r}$ is set to 1 inside the cropped bounding box region and 0 otherwise. The source patch is resized to match $\hat{b}$, and the augmented image $x_i^{\text{aug}}$ is composed as $x_i^{\text{aug}} = M_i^{\text{ins}}\odot x_{s^r}^{\text{src}} +(1-M_i^{\text{ins}})\odot x_i$, where the insertion mask is $M_i^{\text{ins}}(p) = M_{s^r}(p-\hat{b}_{\text{top-left}}) \cdot \mathbf{1}(p\in\hat{b})$. Finally, we update the concept label vector to $o^{\text{aug}}_i$ via $(o^{\text{aug}}_i)_j = \max((o_i)_j, \mathbf{1}(s_j = s^r))$, effectively injecting the missing supervisory signal into the training set while respecting the spatial layout of high-reliability concepts. This procedure is repeated for every rare concept $s^r$, synthesizing new samples and inserting them into $\hat{D}^{\text{tr}}$ until the concept’s frequency exceeds the target threshold. This process forms the augmented dataset $\hat{D}^{\text{tr}}_{\text{aug}}$ and mitigates concept imbalance across the entire dataset.

\begin{figure}[t]
    \centering
    \begin{tikzpicture}
        \node[anchor=south west, inner sep=0] (img) at (0,0)
            {\includegraphics[width=0.9\linewidth]{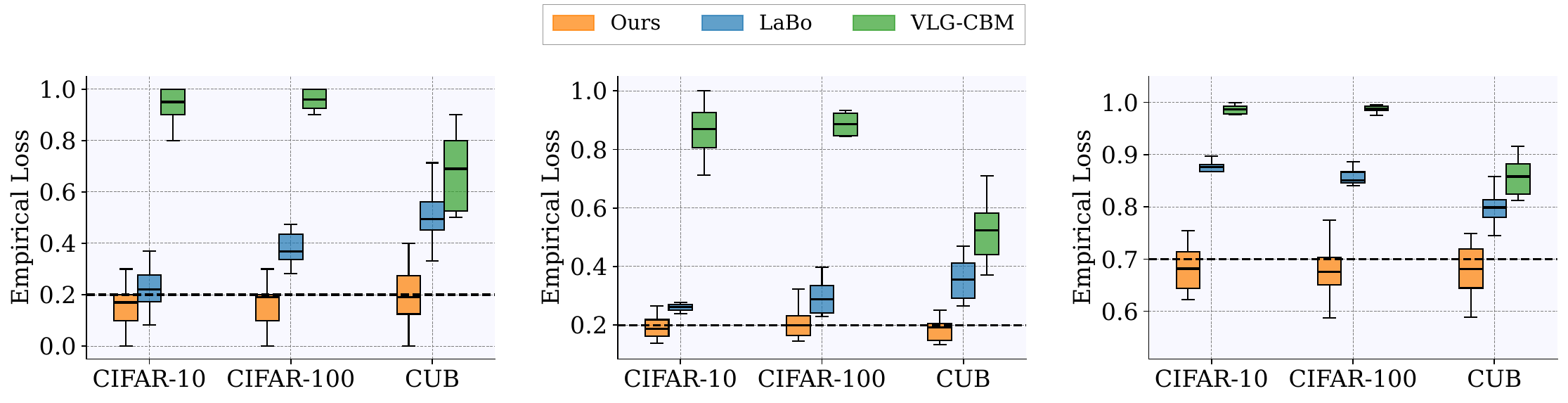}};

        \node at (0.17*0.9\linewidth, -0.02*0.9\linewidth) {(a) Diversity loss};
        \node at (0.513*0.9\linewidth, -0.02*0.9\linewidth) {(b) Coverage loss};
        \node at (0.855*0.9\linewidth, -0.02*0.9\linewidth) {(c) Discriminability loss};
    \end{tikzpicture}
    \vspace{-5mm}
    \caption{Validity comparison of our proposed methods and baselines on CIFAR-10, CIFAR-100, and CUB datasets.}
    \label{fig:validity}
    \vspace{-6.8mm}
\end{figure}
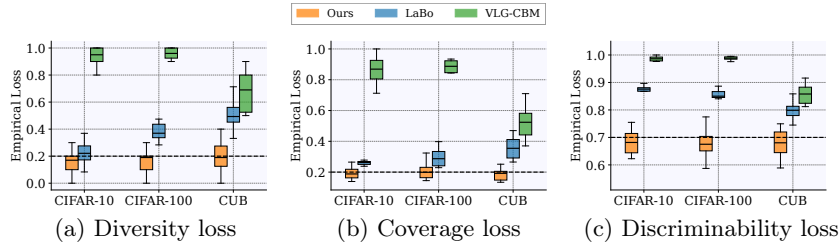

With the constructed concept-labeled dataset $\hat{D}^{\text{tr}}_{\text{aug}}$, we now describe the procedure for training CBMs. Let $\phi: \mathcal{X} \rightarrow \mathbb{R}^d$ be a feature extractor that generates $d$-dimensional embeddings $z_i = \phi(x_i)$ for input image $x_i$. We consider bottleneck models of the form $h(g(\phi(x_i)))$, where $g: \mathbb{R}^d \rightarrow \mathbb{R}^k$ maps an input embedding into the concept space, and $h: \mathbb{R}^k \rightarrow \mathbb{R}^L$ maps concepts into a final class prediction. We call these concept bottleneck models because their prediction $\hat{y}_i = h(g(\phi(x_i)))$ relies on the input $x_i$ entirely through the concept bottleneck $\hat{o}_i = g(\phi(x_i))$, which we train to match our generated concept $o_i$ across all its dimensions.
Our training procedure minimizes a combined loss function composed of two main components. The first is a Binary Cross Entropy (BCE) loss \(L_{C}\), which evaluates the accuracy of the predicted concepts. The second is a Cross Entropy (CE) loss \(L_Y\), measuring performance on the final task prediction. We jointly optimize the predictors $h$ and $g$ as follows:
\vspace{-2.5mm}\begin{align}
\min_{h,g} \mathcal{L}_{C}+\gamma_1\mathcal{L}_{Y}+\gamma_2 R_\beta \ 
\text{s.t.} \ &\mathcal{L}_{C} = \frac{1}{|\hat{D}^{\text{tr}}_{\text{aug}}|} \sum_{(x_i,o_i,y_i)\in \hat{D}^{\text{tr}}_{\text{aug}}}\mathrm{BCE}[g \circ \phi(x_i), o_i ],\\
&\mathcal{L}_{Y} = \frac{1}{|\hat{D}^{\text{tr}}_{\text{aug}}|} \sum_{(x_i,o_i,y_i)\in \hat{D}^{\text{tr}}_{\text{aug}}} \mathrm{CE}[h \circ g \circ \phi(x_i), y_i], \notag
\end{align}
\vspace{-5.6mm}

\noindent where $R_{\beta} = (1 - \beta) \frac{1}{2} \| W_F \|_2^2 + \beta \| W_F \|_1$ is the elastic-net regularization~\cite{zou2005regularization} on weight matrix $W_F$ of the predictor $h$. The hyperparameter  $\gamma_1$ controls the tradeoff between $\mathcal{L}_{C}$ and $\mathcal{L}_{Y}$, and $\gamma_2$ controls the regularization strength. By training on our uncertainty-aware augmented data, we ensure the model effectively captures the informative signals from these concepts, mitigating the impact of sparse supervisory signals caused by varying reliability levels. Note that our method can be easily generalized to the post-hoc model editing setting, where we first apply machine unlearning techniques \cite{chen2025survey,zhao2023static} to remove previously learned deterministic information and then incorporate uncertainty-aware language guidance to update the model, without requiring costly full retraining from scratch~\cite{zhao2024rethinking,qian2023towardsun}.

%% file: secs/5_experiment.tex
\begin{figure}[t]
    \centering
    \includegraphics[width=0.95\textwidth]{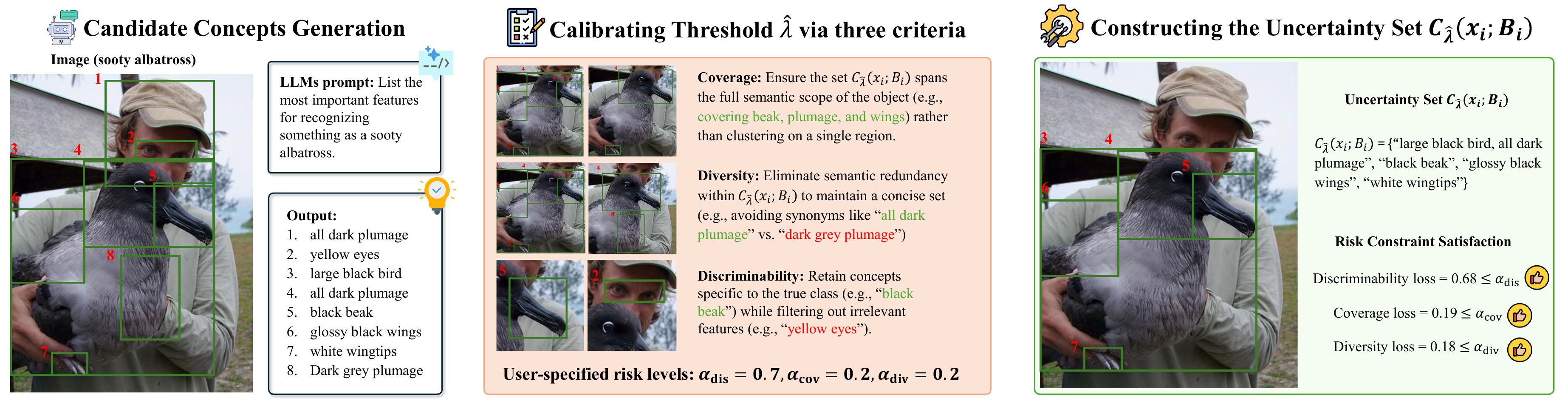}
    \vspace{-4mm}
    \caption{Visualization of the uncertainty-aware concept generation pipeline.}
    \label{fig:visual}\vspace{-6mm}
\end{figure}

\vspace{-4mm}
\section{Experiments}
\label{sec:Exp}
\vspace{-3mm}
\subsection{Experimental Setup }
\vspace{-2mm}
\textbf{Datasets and Models.}
Following prior work, we conduct experiments on three image recognition datasets:
CIFAR-10, CIFAR-100~\cite{krizhevsky2009learning} and CUB~\cite{WahCUB_200_2011}. We generate the initial concept set using GPT-3, following the procedure described in \cite{oikarinen2023label}. To extract the $d$-dimensional embeddings for each input image, we employ the image encoder of the CLIP-RN50 model.

\textbf{Baselines.}  
We compare our method with two major baselines when applicable: LaBo~\cite{yang2023language} and VLG-CBM~\cite{srivastava2024vlg}. LaBo prompts GPT-3 to generate candidate sentence-level concepts, then selects a discriminative, diverse subset and aligns them to images via CLIP. VLG-CBM injects vision-language–grounded object tokens from open-vocabulary detectors to build faithful concept layers.

\textbf{Metrics.}
To verify our theoretical claims, we report the empirical risks on the calibration split and confirm they remain below the desired risk levels in the constraints of Eq.~\eqref{eq:Find}. We also report overall test accuracy and worst-class accuracy to evaluate downstream model performance. Moreover, we introduce a new evaluation metric, Concept Compliance Accuracy (CCA), to quantify how often a test sample is both correctly classified and the concept set used for prediction simultaneously meets all three quality criteria.
Specifically, for each method $m$, we first obtain its instance-level concept predictions for $x_i$ following the method's standard inference procedure, and then construct an effective concept set $\mathcal{C}^{(m)}(x_i)$ by controlling the number of effective concepts (NEC)~\cite{srivastava2024vlg}. Let $\hat{y}_i = h\circ g\circ \phi(x_i)$ denote the predicted label. We then define the constraint indicators
$\delta_{i}^\text{dis} = \mathbf{1}\{\ell_\text{dis}(\mathcal{C}^{(m)}(x_i), x_i, y_i) \leq \alpha_\text{dis}\}$,
$\delta_{i}^\text{cov} = \mathbf{1}\{\ell_\text{cov}(\mathcal{C}^{(m)}(x_i),x_i, y_i) \leq \alpha_\text{cov}\}$,
and
$\delta_{i}^\text{div} = \mathbf{1}\{\ell_\text{div}(\mathcal{C}^{(m)}(x_i),x_i,y_i) \leq \alpha_\text{div}\}$,
along with the task accuracy indicator
$\delta_{i}^* = \mathbf{1}\{\hat{y}_i = y_i\}$.
CCA is then calculated as $\frac{1}{n}\sum_{i=1}^{n}(\delta_{i}^* \land \delta_{i}^\text{dis}\land \delta_{i}^\text{cov} \land \delta_{i}^\text{div})$.

\begin{figure}[t]
    \centering
    \begin{minipage}[t]{0.45\linewidth}
        \centering
        \includegraphics[width=0.9\linewidth]{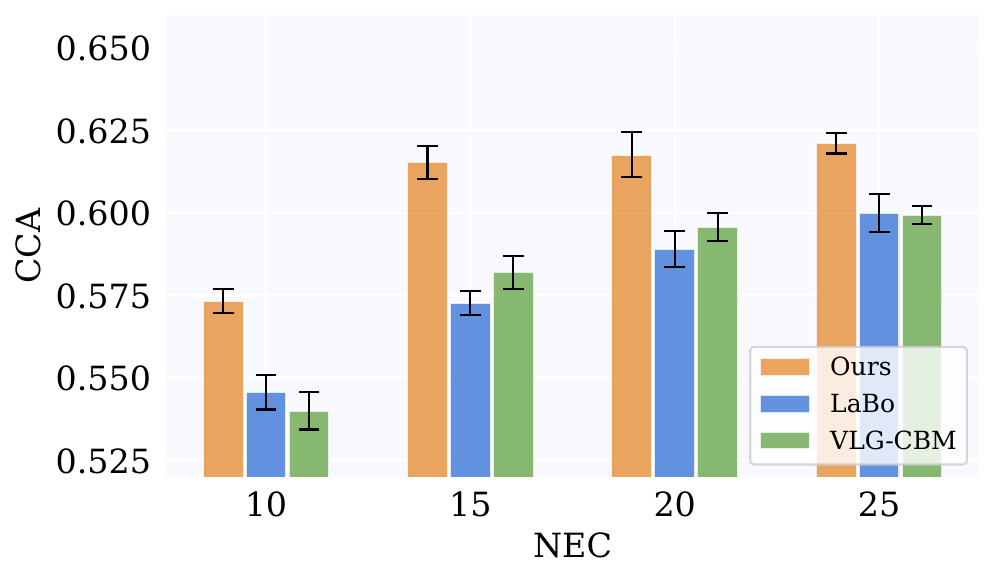}
        \par\vspace{-1mm}
        (a) CIFAR-100 
    \end{minipage}
    \hfill
    \begin{minipage}[t]{0.45\linewidth}
        \centering
        \includegraphics[width=0.9\linewidth]{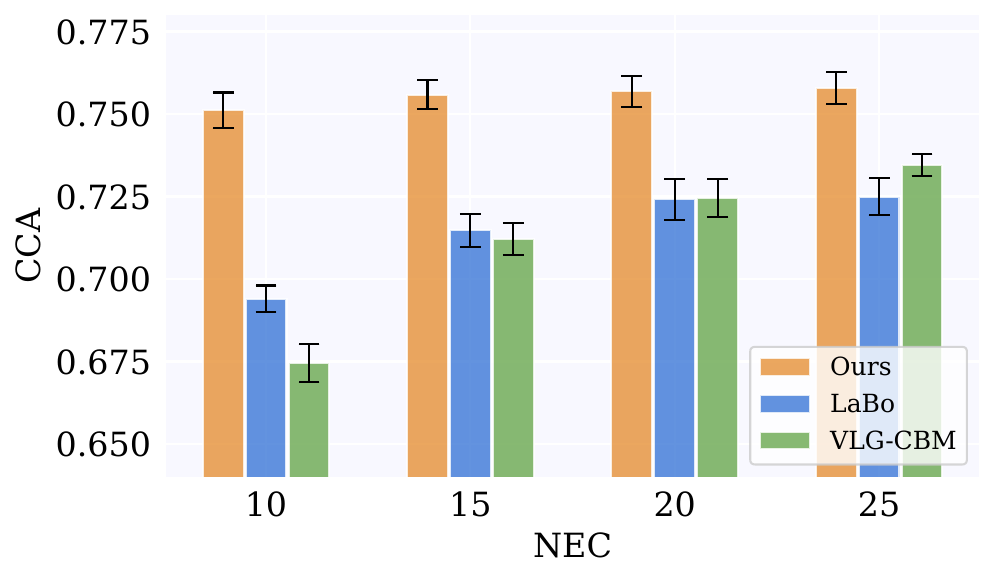}
        \par\vspace{-1mm}
        (b) CUB
    \end{minipage}
    \vspace{-2.3mm}
    \caption{Comparison of concept compliance accuracy on two datasets. }
    \label{fig:cca}
    \vspace{-7mm}
\end{figure}

\textbf{Implementation Details.}
We prompt GPT-3-text-davinci-002 to generate concepts.
We use CLIP-RN50 as the vision backbone, adapted from OpenAI’s public implementation. We use the official test split for evaluation and randomly divide the original training split into 80\% training data and 20\% calibration data. 
All experiments are run for 10 trials, and we report the average results.

\vspace{-4mm}
\subsection{Experimental Results}
\vspace{-2mm}
\textbf{Validity.} We evaluate the validity performance of our approach against the baselines on all three datasets by computing the empirical losses in Eq.~\eqref{eq:Find} at risk levels $\alpha_{\text{div}}=0.2$, $\alpha_{\text{cov}}=0.2$, and $\alpha_{\text{dis}}=0.7$. Fig.~\ref{fig:validity} displays boxplots for each loss: the dashed line denotes the desired risk level, and the horizontal bar inside each box marks the mean empirical loss. Across all datasets, each loss achieved by our methods stays at or below its corresponding $\alpha$, thereby satisfying the constraints in 
Eq.~\eqref{eq:Find}. As an example, for the desired risk level $\alpha_{\text{dis}}=0.7$ on CIFAR-10, our method achieves a discriminability loss of $0.68$, successfully satisfying the constraint. In contrast, VLG-CBM and LaBo record losses of 0.99 and 0.88, respectively, both failing to meet this $\alpha_{\text{dis}}=0.7$ threshold. These results demonstrate our method's unique ability to provide formally guaranteed uncertainty control for the generated concepts. To provide qualitative insight, Fig.~\ref{fig:visual} visualizes the uncertainty quantification pipeline on a sample image. We observe that the raw candidate concepts generated by LLMs often contain hallucinations (e.g., yellow eyes) and semantic redundancies. Our method successfully filters these out using the derived threshold $\hat{\lambda}$: the discriminability criterion rejects the incorrect ``yellow eyes'', the diversity criterion eliminates redundant concept ``dark grey plumage'', and the coverage criterion ensures the final set spans the full semantic scope of the object. The resulting concept uncertainty set not only aligns with human perception but also empirically satisfies all risk constraints, consistent with the quantitative guarantees demonstrated in Fig.~\ref{fig:validity}.

\begin{figure}[t]
    \centering
    \begin{minipage}[t]{0.45\linewidth}
        \centering
        \includegraphics[width=0.9\linewidth]{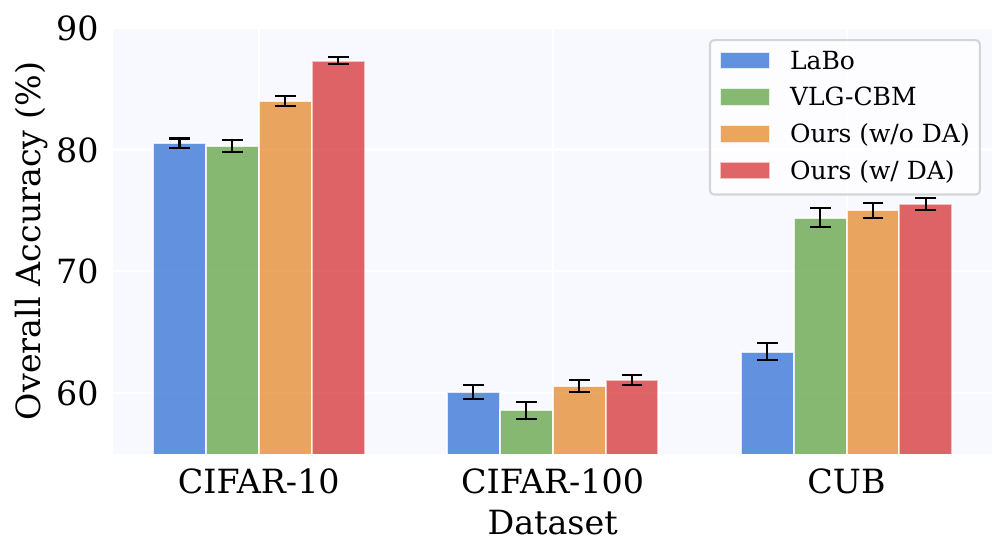}
        \par\vspace{-1mm}
        (a) Overall accuracy 
    \end{minipage}
    \hfill
    \begin{minipage}[t]{0.45\linewidth}
        \centering
        \includegraphics[width=0.9\linewidth]{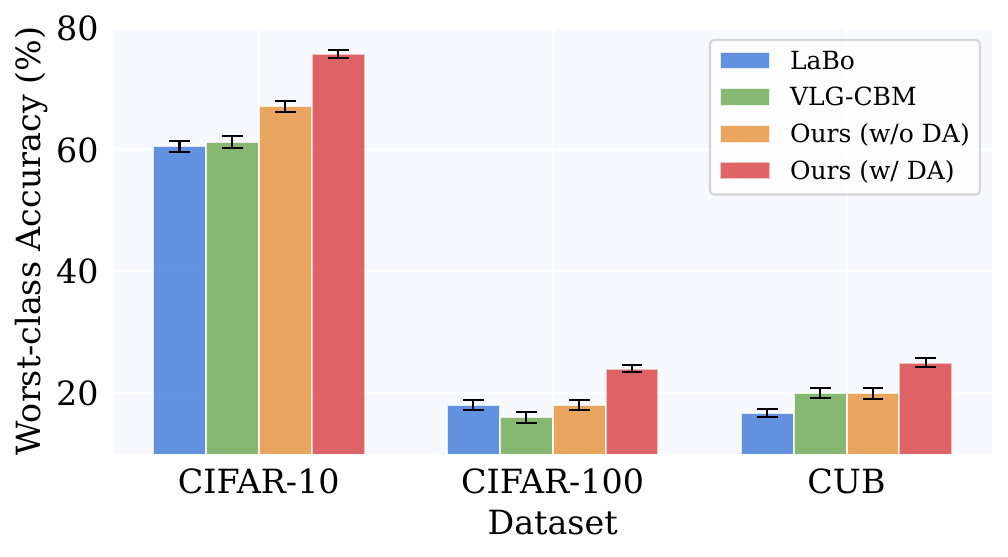}
        \par\vspace{-1mm}
        (b) Worst-class accuracy
    \end{minipage}
    \vspace{-2.5mm}
    \caption{Test accuracy comparison on CIFAR-10, CIFAR-100, and CUB datasets.}
    \label{fig:acc}
    \vspace{-7mm}
\end{figure}

\textbf{Concept Compliance Accuracy.}
Beyond individual risk control, we evaluate how well our framework produces concept sets that simultaneously satisfy all quality constraints while leading to a correct final prediction. Fig.~\ref{fig:cca} shows the CCA of our methods and baselines as NEC increases from 10 to 25. Across this entire NEC range on both CIFAR-100 and CUB, our method consistently attains the highest CCA, indicating that a larger share of test samples meet all three risk constraints while being correctly classified. For example, on CUB, our CCA starts at 0.75 when NEC\(=\)10, versus 0.67 for VLG‑CBM and 0.69 for LaBo. 
These results demonstrate that, regardless of how many effective concepts are retained, our approach ensures a larger number of test samples whose concept sets both satisfy the constraints in Eq.~\eqref{eq:Find} and lead to correct predictions, outperforming the baselines across the board.

\textbf{Test Accuracy.} We compare the overall and worst-class test accuracy across CIFAR‑10, CIFAR‑100, and CUB in Fig.~\ref{fig:acc}. Our methods achieve the strongest performance across both metrics.
In the overall accuracy metric, our model trained with data augmentation reaches 75.5\% on CUB, exceeding VLG-CBM at 74.4\% and LaBo at 63.4\%. This superior performance is primarily due to our uncertainty-aware concept selection, which, based on formal guarantees, produces a more reliable and effective concept vocabulary by filtering out flawed annotations.
Furthermore, the advantage is even more pronounced in the worst-class accuracy metric. For example, our model with data augmentation achieves 25.0\% on CUB. This surpasses LaBo at 16.7\% and VLG-CBM at 20.0\%, and significantly improves upon our own model without data augmentation, which scored 20.0\%. This clear boost stems from our targeted data augmentation pipeline guided by uncertainty metrics, which mitigates the scarcity of valid training samples and ensures the model effectively incorporates the informative signals from sparse concepts that would otherwise be disregarded.

%% file: secs/6_conclusion.tex
\vspace{-4.5mm}
\section{Conclusion}
\label{sec:Conclusion}
\vspace{-3mm}

In this paper, we propose a principled framework for uncertainty-aware CBMs guided by LLMs, characterized by rigorous uncertainty quantification with formal guarantees and an uncertainty-aware training strategy that effectively leverages concepts of varying reliability. Specifically, we establish a principled concept selection mechanism grounded in CP, which calibrates uncertainty sets across discriminability, coverage, and diversity criteria to provide theoretical guarantees. This mechanism successfully addresses the primary challenge of providing formal quality guarantees for LLM-annotated concepts without relying on strict distributional assumptions. Furthermore, we develop a targeted data augmentation pipeline that leverages our derived concept uncertainty to guide the synthesis of valid training samples. This approach effectively resolves the critical challenge where varying reliability levels lead to sparse supervisory signals, thereby alleviating the scarcity of reliable concept annotations and ensuring the model effectively utilizes these informative signals. We also provide the theoretical analysis for our proposed methods. Extensive experiments on real-world datasets demonstrate the advantages of the proposed mechanism.

\vspace{-4mm}
\subsubsection{\ackname} This work is supported in part by the US National Science Foundation under grants CNS-2350332 and IIS-2442750. Any opinions, findings, and conclusions or recommendations expressed in this material are those of the author(s) and do not necessarily reflect the views of the National Science Foundation.